\documentclass[conference]{IEEEtran}
\IEEEoverridecommandlockouts

\usepackage{cite}
\usepackage{amsmath,amssymb,amsfonts}
\usepackage{algorithmic}
\usepackage{graphicx}
\usepackage{textcomp}
\usepackage{xcolor}

\usepackage{subfigure}
\usepackage{multirow}
\usepackage{makecell}
\usepackage{latexsym}
\usepackage{amsthm}
\usepackage{booktabs}
\usepackage{enumitem}
\usepackage[normalem]{ulem}  %
\usepackage[ruled,vlined,linesnumbered]{algorithm2e}
\usepackage{tabularx}
\usepackage{tikz}
\usepackage{eso-pic}

\def\BibTeX{{\rm B\kern-.05em{\sc i\kern-.025em b}\kern-.08em
    T\kern-.1667em\lower.7ex\hbox{E}\kern-.125emX}}
\begin{document}

 \AddToShipoutPictureBG*{%
  \AtPageUpperLeft{%
    \raisebox{-1.5cm}{%
      \parbox{\paperwidth}{%
        \centering
        \small\itshape
        This work has been submitted to the IEEE for possible publication. \\
        Copyright may be transferred without notice, after which this version 
        may no longer be accessible.
      }%
    }%
  }%
}

\title{Robust Energy-Aware Routing for Air-Ground Cooperative Multi-UAV Delivery in Wind-Uncertain Environments
\\
}


\author{Tianshun Li$^{1}$, Hongliang Lu$^{2}$, Yanggang Sheng$^{1}$, Zhongzhen Wang$^{1}$, Haoang Li$^{3}$ and Xinhu Zheng$^{4, \ast}$ 
\thanks{$^\ast$ Corresponding author.}
\thanks{$^{1}$Tianshun Li, Yanggang Sheng, and Zhongzhen Wang are with
        The Hong Kong University of Science and Technology (Guangzhou), China
        {\tt\small tli449@connect.hkust-gz.edu.cn, yanggangs@hkust-gz.edu.cn, wzz1011@hotmail.com}}%
        \thanks{$^{2}$Hongliang Lu is with The Hong Kong University of Science and Technology, HongKong
        {\tt\small honglianglu@ust.hk}}
        \thanks{$^{3}$Haoang Li is with the Intelligent Transportation Thrust and Robotics and Autonomous Systems Thrust, Systems Hub, The Hong Kong University of Science and Technology (Guangzhou), China
        {\tt\small haoangli@hkust-gz.edu.cn}}
        \thanks{$^{4}$Xinhu Zheng is with the Intelligent Transportation Thrust, Systems Hub, Internet of Things Thrust, Information Hub, The Hong Kong University of Science and Technology (Guangzhou), China
        {\tt\small xinhuzheng@hkust-gz.edu.cn}}%
}

\maketitle

\begin{abstract}
Ensuring energy feasibility under wind uncertainty is critical for the safety and reliability of UAV delivery missions. In realistic truck–drone logistics systems, UAVs must deliver parcels and safely return under time-varying wind conditions that are only partially observable during flight. However, most existing routing approaches assume static or deterministic energy models, making them unreliable in dynamic wind environments.
We propose Battery-Efficient Routing (BER), an online risk-sensitive planning framework for wind-sensitive truck-assisted UAV delivery. The problem is formulated as routing on a time-dependent energy graph whose edge costs evolve according to wind-induced aerodynamic effects. BER continuously evaluates return feasibility while balancing instantaneous energy expenditure and uncertainty-aware risk. The approach is embedded in a hierarchical aerial–ground delivery architecture that combines task allocation, routing, and decentralized trajectory execution. Extensive simulations on synthetic ER graphs generated in Unreal Engine environments and quasi-real wind logs demonstrate that BER significantly improves mission success rates and reduces wind-induced failures compared with static and greedy baselines. These results highlight the importance of integrating real-time energy budgeting and environmental awareness for UAV delivery planning under dynamic wind conditions.
\end{abstract}

\section{Introduction}

UAVs are increasingly applied in civilian fields such as agriculture, monitoring, and urban logistics, with last-mile delivery becoming a key use case  \cite{Hu2025MultiDroneTruckCD,Park2024LearningBasedCM}. Compared with ground transport, UAVs provide faster access and greater flexibility \cite{Xu2025UAVAT}. However, the limited onboard battery capacity restricts the operational range of standalone UAV delivery. 
To extend the delivery range and improve efficiency, truck-assisted drone systems have emerged \cite{2025Multi}. In this model, a single drone is transported by a truck and can launch from or recover to either the truck or a depot \cite{Murray2015TheFS}.  
While truck-assisted systems alleviate geometric range limitations, they do not fundamentally resolve energy feasibility under environmental uncertainty, particularly wind, a dominant factor governing mission feasibility.

\begin{figure}
    \centering
    \includegraphics[width=1\linewidth]{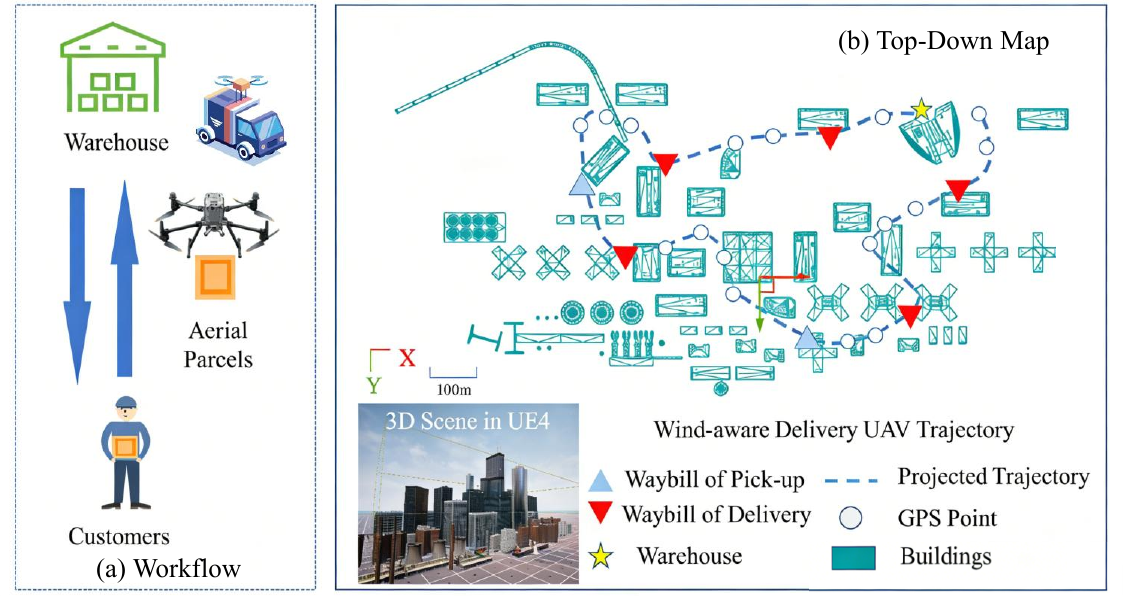}
    \caption{Wind-sensitive energy-efficient routing and stage-wise energy allocation under dynamic flow conditions.
(a) Workflow of truck-UAV system.
(b) Simulated environment of the delivery system. UAVs depart on a moving truck from a warehouse, transports a parcel to a designated customer, and must retain sufficient residual energy to guarantee a safe return flight.
}
    \label{fig:Illustration}
\end{figure}

Among environmental disturbances, wind directly alters propulsion energy consumption, thereby tightening the effective energy budget available for the safe return of UAVs \cite{chodnicki2022energy}. Specifically, energy consumption is direction-dependent and time-varying because wind directly modulates propulsion demand by altering relative airspeed. Under constant airspeed, headwind increases required thrust and power consumption, while tailwind reduces energy use. Since energy depends on both distance and wind conditions \cite{meier2022wind}, mission feasibility is dynamic: a path initially feasible may become unsafe under changing wind. Therefore, ensuring safe return under wind uncertainty is critical for UAV delivery systems. 

Moreover, UAVs rarely operate in isolation in real-world urban logistics. Commercial deployments typically involve multiple drones dispatched from a shared mobile depot to serve spatially distributed customers. 
Fig. \ref{fig:Illustration} shows such a truck-assisted multi-drone system, with a mobile truck acting as a shared mobile depot. Each UAV performs single-customer deliveries, and urban wind fields are spatially heterogeneous, causing asymmetric energy costs. Furthermore, payload influences thrust demand \cite{Bi2024TruckDroneDO}.
Existing routing approaches typically assume static or known costs, modeling wind as constant or offline information \cite{Wu2021ReinforcementLB}. However, real wind is time-varying and locally observable, making routing an online decision problem. 
This mismatch between static assumptions and dynamic wind motivates an online feasibility reassessment mechanism.
Static shortest paths may become infeasible mid-mission \cite{duan2024energy}. As a consequence, routes computed offline may violate safety margins during flight, potentially preventing a safe return.

More than trajectory-level feasibility, real-world UAV deployments introduce an additional layer of complex interdependencies across multiple UAVs and the mobile truck.
In multi-drone-truck systems, truck mobility reshapes feasible return regions, and assignment decisions affect energy safety margins \cite{chiang2019impact,wang2019overview}. Directly solving the joint routing and scheduling problem in a centralized manner leads to a high computational burden due to truck relocation, and wind-dependent path costs. This coupling between routing, assignment, truck relocation, and wind uncertainty renders centralized optimization computationally intractable.

To tackle these challenges, we design a hierarchical decision architecture. 
At the system level, task allocation decomposes the large-scale delivery 
problem into tractable subproblems. At the routing level, we introduce 
Battery-Efficient Routing (BER), which performs wind-sensitive online path 
planning under energy constraints. Finally, decentralized reinforcement learning controllers 
are used for local trajectory execution and collision avoidance among 
multiple drones.
Our contributions can be summarized as:

1) 
We introduce a graph representation where edge weights evolve based on wind-dependent aerodynamic power modeling. This model allows for a formal definition of mission feasibility that incorporates battery return constraints directly into the routing framework.

2) 
We propose the Battery-Efficient Routing (BER) framework, a modular energy-aware planning approach that explicitly safeguards mission feasibility under wind uncertainty. 

3)
We conduct comparative experiments, demonstrating that our approach exhibits significant performance advantages over existing baselines methods across various scenarios in both sustainability and the number of in-route drone collisions.

\section{Related Works}

\subsection{Energy-Aware UAV Planning}

Energy-aware UAV routing aims to extend flight endurance and improve delivery reliability under strict onboard power constraints. Early studies primarily formulated trajectory design as an optimization or decision-making problem in which energy consumption depended on flight dynamics, path length, and smoothness \cite{Deng2025TargetAA}. Mellinger and Kumar~\cite{mellinger2012trajectory} proposed differential-flatness-based trajectory generation to reduce aggressive maneuvers and load oscillations. Later, MDP-based frameworks modeled energy accumulation explicitly and optimized long-horizon control policies under stochastic dynamics~\cite{han2020energy,michel2023energy}. These works established the importance of integrating vehicle dynamics into planning but often relied on simplified environments or stationary cost assumptions.

More recent research incorporates energy models or multi-agent reinforcement learning into drone delivery and path planning systems \cite{Hu2025MultiDroneTruckCD}. Dorling et al.~\cite{dorling2016vehicle} propose a vehicle routing formulation for drones that integrates energy limits into combinatorial allocation. Subsequent approaches combine obstacle avoidance and path smoothing with battery-aware cost modeling \cite{9766183}, and some frameworks now incorporate aerodynamic factors into flight-time estimation \cite{Pasha2026EnablingTM}. However, most existing formulations assume that environmental conditions remain constant during mission execution. Wind, when considered, is typically treated as a static disturbance or average correction factor.

\subsection{Routing on Dynamic and Time-Dependent Graphs}

Shortest-path planning on weighted graphs is a classical problem~\cite{wang2024review}. When edge costs are fixed and globally known, Dijkstra-type algorithms provide efficient optimal solutions. Later research extended these algorithms to dynamic graphs where edge weights change over time~\cite{zaki2016comprehensive}. Incremental shortest-path methods update the solution tree when costs are externally modified~\cite{rasmussen2008tree}.

Despite these advances, dynamic graph algorithms typically assume that weight changes are observable and can be incorporated deterministically~\cite{du2023efficient}. In UAV routing under wind uncertainty, this assumption no longer holds. In \cite{ferone2017shortest,yin2021learning}, a new dynamic routing algorithm is proposed using all possible shortest paths to quickly compute the new shortest paths in case of edge additions or deletions. To the best of our knowledge, the online approach for the shortest path problem has been pursued mainly for the stochastic shortest path problem. Such a problem has been long studied in the machine learning community, for example, in the framework of adversarial bandit problem \cite{2024Energy,zhu2021uav} using a Markov decision problem where an agent moves in an acyclic graph with random transitions. This context is completely different from ours: the reward seems local, the graph is acyclic, and the weights are stochastic.

\section{Problem Formulation}

\subsection{The Relative Wind Model}

\begin{figure}
    \centering
    \includegraphics[width=1\linewidth]{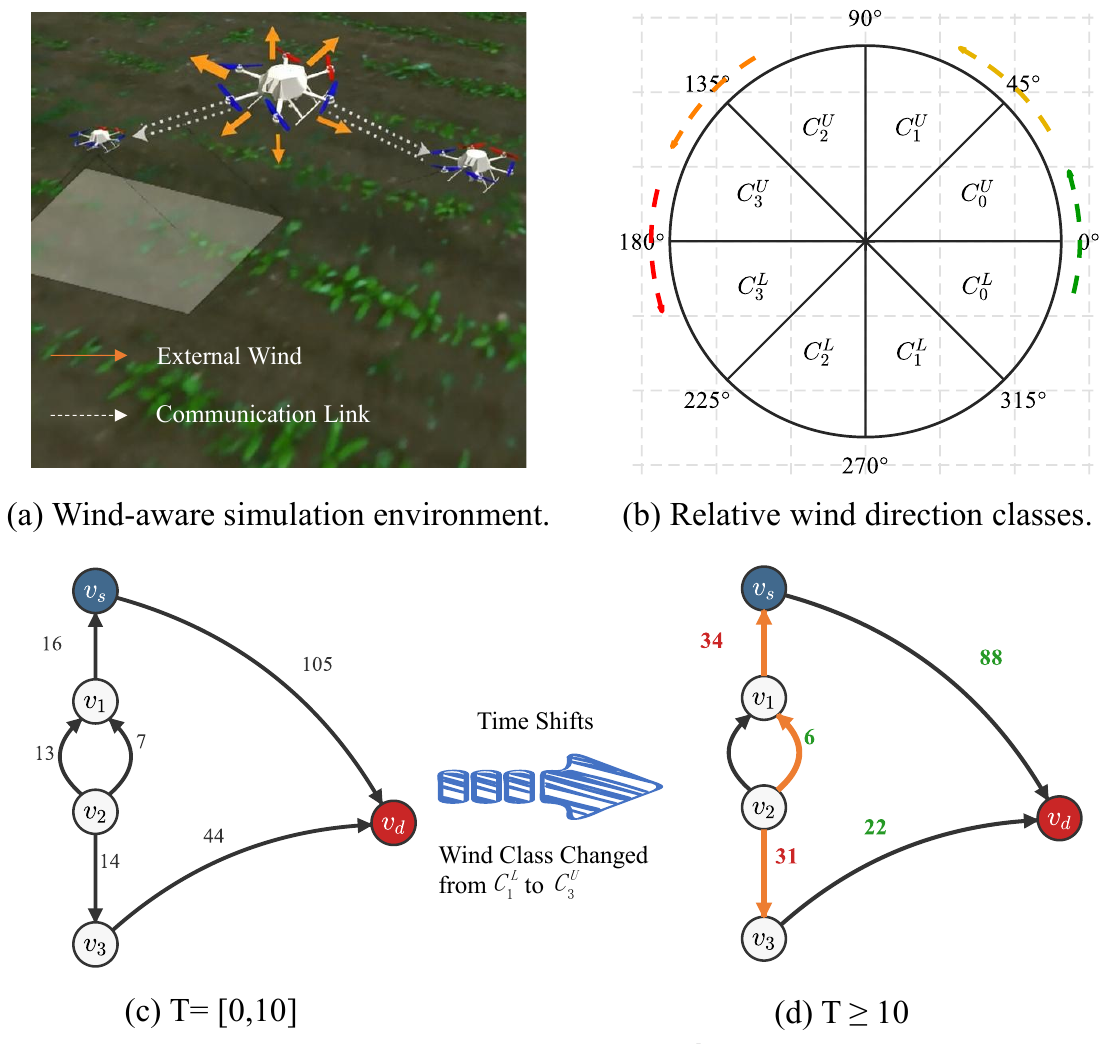}
    \caption{Wind-sensitive modeling and time-dependent edge weight variation.
(a-b) represent wind disturbance visualization and relative wind direction discretization.
(c-d) denote time-window-based edge cost updates under wind class transitions.}
    \label{fig:hebing}
\end{figure}

As shown in Fig. \ref{fig:hebing}, we model the UAV motion as a Dubins path relative to the ground. 
We consider the state space of the vehicle as 
$\mathbf{x} = (x, y, z, \theta^G)$, 
where $(x,y,z)$ denotes the three-dimensional position and $\theta^G$ is the ground-relative heading angle. 
To capture the differential-flatness-based kinematic approximation of quadrotors for path planning, we adopt a simplified Dubins-like model with curvature constraints~\cite{4434966}.
The kinematic constraints of path $\boldsymbol\eta$ are expressed as:

\begin{align}
\frac{\partial \boldsymbol{\eta}}{\partial s}
= f(\boldsymbol{\eta}(s))
=
\begin{pmatrix}
\cos(\gamma^G)\cos(\theta^G) \\
\cos(\gamma^G)\sin(\theta^G) \\
\sin(\gamma^G) \\
\kappa \cos(\gamma^G)
\end{pmatrix},
\label{eq:dubins_airplane}
\end{align}
where \(s\) is the arc length parameter, $\gamma^G$ denotes the flight-path angle and $\kappa$ the curvature. 
The vehicle maneuverability is bounded by:
\begin{equation}
    \gamma^G \in [\gamma_{\min}^G, \gamma_{\max}^G], 
\quad 
\kappa \in [-\kappa_{\max}, \kappa_{\max}].
\end{equation}

Given the planned path (Eq.~\ref{eq:dubins_airplane}), the air-relative velocity $\mathbf{V}^A$ is computed such that the resulting ground velocity $\mathbf{V}^G$ aligns with the path direction $\mathbf{u}_\eta$. 

Since aerial vehicles generate forces in an air-relative medium, 
the kinematics are formulated in the air-relative frame. 
The ground velocity $\mathbf{V}^G$ follows the wind triangle relation (Fig.~2):
\begin{align}
\mathbf{V}^G = \mathbf{V}^A + \mathbf{W}.
\label{eq:wind_triangle}
\end{align}
where $\mathbf{V}^A = [V_x^A, V_y^A, V_z^A]^\top$ denotes the air-relative velocity, $\mathbf{W}$ denotes wind velocity. 
Assuming UAVs operate at an efficiency-optimal airspeed, 
we consider a constant magnitude:
\begin{equation}
    V^A = \|\mathbf{V}^A\|.
\end{equation}

The air-relative heading $\theta$ and flight-path angle $\gamma$ are defined as:
\begin{align}
\theta^A &= \operatorname{atan2}(V_y^A, V_x^A), 
& \theta^A &\in [0, 2\pi], \label{eq:thetaA} \\
\gamma^A &= \arcsin\!\left(\frac{V_z^A}{V^A}\right), 
& \gamma^A &\in [\gamma_{\min}^A, \gamma_{\max}^A]. \label{eq:gammaA}
\end{align}

Note that maneuverability limits, particularly the flight-path angle constraint, 
are intrinsically defined in the air-relative frame.
Decomposing Eq.~\ref{eq:wind_triangle} into tangential ($\parallel$) and normal ($\perp$) components yields:
\begin{align}
V^G \mathbf{u}_\eta &= \mathbf{V}^A + \mathbf{W}_{\parallel}, \label{eq:tangent_iros} \\
0 &= \mathbf{V}^A_{\perp} + \mathbf{W}_{\perp}. \label{eq:normal_iros}
\end{align}

Assuming constant airspeed magnitude $V^A$, the tangential airspeed component and ground speed are:
\begin{align}
V^A_{\parallel} &= \pm \sqrt{(V^A)^2 - (W_{\perp})^2}, \label{eq:VA_para_iros} \\
V^G &= V^A_{\parallel} + W_{\parallel}. \label{eq:VG_iros}
\end{align}

A path is infeasible if either of the following holds:
\begin{itemize}
    \item $|W_{\perp}| > V^A$, implying no real solution for $V^A_{\parallel}$;
    \item $V^G \le 0$, indicating that the path direction is unreachable under the given wind condition.
\end{itemize}

\subsection{Energy Consumption Model}
We model the delivery environment as a directed graph $G=(V,E)$, where each edge $e\in E$ corresponds to a feasible flight segment of length $\ell(e)$. 
We adopt a simplified longitudinal equilibrium model under the following assumptions:
(i) small angle of attack such that thrust $T$ aligns with airspeed,
(ii) constant drag $D$ due to constant airspeed operation.
Under these assumptions, the aircraft dynamics reduce to:

\begin{align}
0 = T - D - mg \sin(\gamma^A),
\label{eq:long_equilibrium}
\end{align}
where $m$ is the vehicle mass and $g$ the gravitational constant.
The required thrust $T$ is therefore approximated as:

\begin{align}
T \approx \max\!\left(D + mg \sin(\gamma^A),\, 0 \right).
\label{eq:thrust_model}
\end{align}

For descending flight, $D + mg \sin(\gamma^A)$ may become negative,
indicating acceleration. Since we assume static equilibrium,
this yields a conservative energy estimate by neglecting potential
energy recovery.

We define the power consumption model as:

\begin{align}
P = P_c + \frac{T V^A}{c_T},
\label{eq:power_model}
\end{align}
where $P_c$ denotes constant avionics power and $c_T$ is the thrust power coefficient. 
More detailed propulsion models can be incorporated without modifying the planner structure.

Based on the above wind-sensitive kinematic and energy models, we now develop a 
planning framework capable of dynamically adjusting UAV routes under 
partially observable wind conditions.

\section{Method}
The proposed framework follows a two-layer architecture, illustrated in Fig.~\ref{fig:framework}.
At the upper level, task allocation and clustering decompose
the multi-UAV–truck system into several service areas.
At the lower level, BER serves as the core wind-sensitive
energy routing module operating on a time-dependent
energy graph. The reward function is employed only for decentralized
local trajectory execution and collision avoidance.

\subsection{LLM Clustering}
\begin{figure}
    \centering
    \includegraphics[width=1\linewidth]{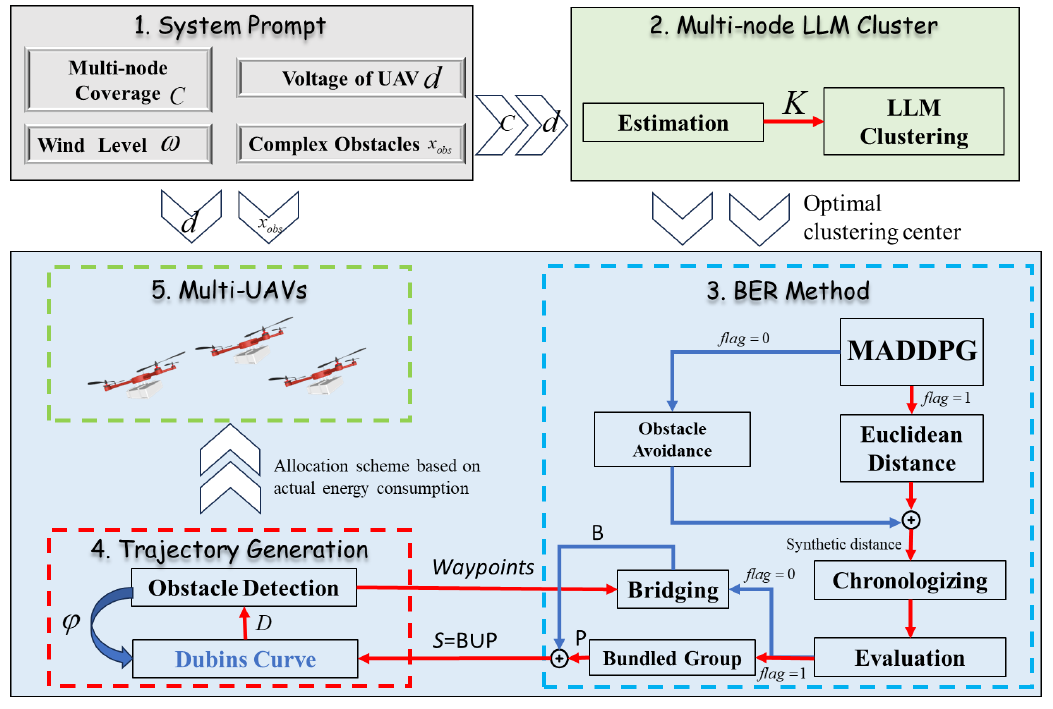}
    \caption{Overview of the proposed framework. System-level constraints (coverage, wind level, battery voltage, and obstacle complexity) condition multi-node LLM clustering.}
    \label{fig:framework}
\end{figure}

In the structured system prompt, as shown in Module~1 of Fig.~\ref{fig:framework}, four key factors are considered: multi-node coverage $C$, wind level $\omega$, UAV voltage $d$, and obstacle complexity $x_{\text{obs}}$. These variables characterize environmental dynamics, energy availability, and geometric constraints.

We then introduce a lightweight LLM-conditioned clustering module that
partitions customers according to energy and wind context.
Customers are first partitioned into truck-served $c$ and drone-served sets $d$.
A truck route is then generated by solving a TSP over the truck-served nodes \cite{tong2022optimal}.
Subsequently, drone assignments are determined based on the obtained truck route.

The task allocation and scheduling process is summarized as follows:

\begin{enumerate}
    \item \textbf{Initialization:}
    At mission start, the depot and customer locations,
    number of drones, parcel weights, vehicle specifications,
    and environmental conditions are retrieved from the simulation environment.
    These inputs are used to compute the truck route via a heuristic TSP solver.

    \item \textbf{Dynamic Assignment:}
    At regular intervals or upon truck arrival,
    the upper solver queries currently available trucks and drones,
    and filters feasible customers based on range, payload capacity,
    and service type constraints.
    Among the feasible customers, the nearest to the truck is assigned
    to the fastest available drone.
    If truck service is required, the next customer on the precomputed
    truck route is assigned to the truck.
\end{enumerate}

\begin{figure}
    \centering
    \includegraphics[width=1\linewidth]{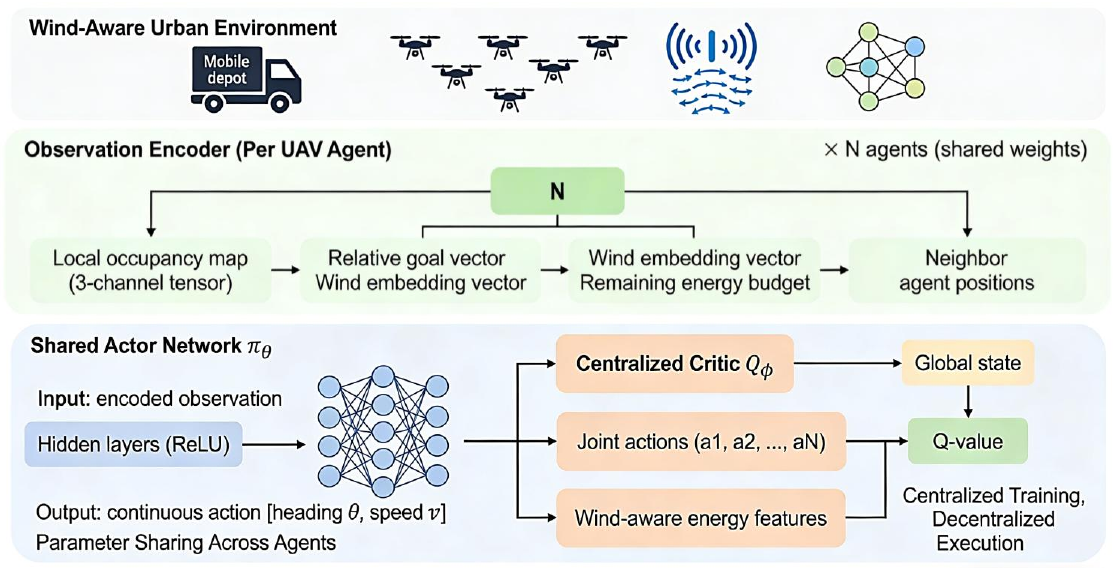}
    \caption{Neural network architecture of the MADDPG model.}
    \label{fig:Network}
\end{figure}

\subsection{BER Core}
Intuitively, BER performs online routing while continuously checking whether the remaining energy budget is sufficient to complete a safe \emph{deliver-and-return} mission under the current wind estimate. At each decision step, the planner selects the next flight segment by jointly considering energy efficiency and wind-induced uncertainty, while pruning actions that would violate the return-feasibility constraint.

As illustrated in Module~3 of Fig.~\ref{fig:framework}, routing is performed on a directed waypoint graph $G=(V,E)$, where each vertex $v\in V$ denotes a feasible waypoint and each edge $e=(v,u)\in E$ represents a flyable segment. 
To incorporate wind influence into path planning, each edge is assigned a wind-sensitive synthetic distance:
\begin{equation}
    L_{SE} = L_{\text{geo}} + \lambda L_{\text{wind}},
    \label{eq:edge}
\end{equation}
where $L_{\text{geo}}$ is the Euclidean distance between waypoints and $L_{\text{wind}}$ captures the additional energy penalty induced by the relative wind. 
The coefficient $\lambda \ge 0$ controls the trade-off between geometric efficiency and wind-aware risk sensitivity.

Let $B$ denote the initial energy budget of the UAV, $v_0$ the depot, and $v_t$ the delivery target. 
At time step $t$, the UAV observes the local wind $w_t$ at its current vertex and updates an online wind estimate $\hat{w}_t$ under partial observability. 
For each outgoing edge $e$, BER evaluates a risk-aware surrogate cost:
\begin{equation}
\tilde c_t(e)=\textsc{EnergyCost}(e,\hat{w}_t)+\lambda\,\textsc{Unc}(e,\hat{w}_t),
\end{equation}
where $\textsc{EnergyCost}(\cdot)$ predicts the energy consumption under the estimated wind $\hat{w}_t$, and $\textsc{Unc}(\cdot)$ quantifies wind-induced uncertainty. 

To guarantee mission safety, BER further invokes a return-feasibility check $\textsc{ReturnOK}(\cdot)$ that ensures the remaining budget can still support a conservative return to the depot. 
A curve factor $\kappa>1$ is introduced to provide a safety margin for the return path. 
Among feasible actions, the function $\textsc{Score}(\cdot)$ ranks candidates according to a threshold-based policy $\pi_{\mathrm{th}}$, and the process continues until the delivery is completed or the maximum number of decision steps $T_{\max}$ is reached.

For details on the Multi-Agent Deep Deterministic Policy Gradient (MADDPG) \cite{11214461} applied in trajectory execution:
\paragraph{ Agent Model}
Each delivery drone is modeled as an independent agent within the environment. 
We consider the same drone types in terms of flight range, payload capacity, and maximum cruising velocity.

\paragraph{ Observation Space}
We model a partially observable urban delivery environment consisting of a depot (truck), customer locations, and high-rise buildings acting as obstacles. 
Each drone has access only to local environmental information within a limited field-of-view (FOV) centered at its current position, consistent with realistic onboard sensing constraints. 
The FOV size is set to $155 \times 155$ in the simulation. 

As shown in Fig. \ref{fig:Network}, the observation representation follows a structured encoding inspired by prior reinforcement-learning-based path planning frameworks, but adapted to our heterogeneous truck-drone setting. 
The observation consists of two components.

\textit{(i) Local spatial tensor:}
A binary tensor encoding the environment within the agent's FOV, organized into three channels:
\begin{itemize}
    \item obstacle occupancy,
    \item neighboring drone positions,
    \item goal locations of neighboring drones.
\end{itemize}

\textit{(ii) Global goal vectors:}
To compensate for partial observability, two auxiliary vectors are provided:
\begin{itemize}
    \item the relative vector from the agent to its current goal,
    \item the relative vector from the truck to its designated goal.
\end{itemize}
The second vector becomes a zero vector when the agent's objective is direct customer delivery rather than returning to the truck.

\paragraph{Action Space}
We assume each drone can move in an arbitrary direction with a speed bounded by its maximum velocity. 
Accordingly, the action space is defined as a normalized two-dimensional continuous vector:

\begin{equation}
A_i = [\theta, v_d],
\end{equation}
where $\theta$ denotes the heading angle and $v_d$ the normalized speed command. 
Both variables are scaled to $[-1,1]$. 
At each time step, the executed motion is determined by a lower-level controller based on the selected action.

\paragraph{Reward Structure}
Within the POMDP framework \cite{Lauri_2023}, the objective of each agent is to complete the assigned delivery task and return to the truck for replenishment while minimizing travel time and avoiding collisions. To maintain consistency between global routing and local control, the wind-induced energy cost $L_{\text{wind}}$
 is incorporated into the reward function as a penalty term weighted by the same risk parameter $\lambda$.
The total reward is formulated as:

\begin{equation}
r = r_{\mathrm{urgency}} + r_{\mathrm{goal}}  
+ r_{\mathrm{coll\_obst}} + r_{\mathrm{coll\_drone}} -  \lambda L_{\text{wind}}.
\end{equation}

\textbf{Time penalty.} 
A step-wise penalty $r_{\mathrm{urgency}} = -0.2$ is imposed at every time step, corresponding to approximately two seconds of real-world flight, to discourage inefficient trajectories and encourage timely task completion.

\textbf{Goal reward.}
A terminal reward $r_{\mathrm{goal}} = 20$ is granted when the Euclidean distance between the drone and its current goal (customer or truck) falls below 20 meters.
To alleviate sparse reward issues, we introduce a distance-based shaping term:

\begin{equation}
r_{\mathrm{near}}(t) = (d_t - d_{t+1}) \cdot k_1,
\end{equation}
where $d_t$ and $d_{t+1}$ denote the distances to the goal at time steps $t$ and $t+1$, respectively, and $k_1 = 0.02$ is a scaling coefficient. 
Positive reward is obtained when the agent moves closer to the goal.

\textbf{Obstacle collision penalty.}
A penalty $r_{\mathrm{coll\_obst}} = -2$ is imposed upon collision with static obstacles.
To promote safe multi-agent coordination, we incorporate a proximity-based penalty inspired by potential field formulations:

\begin{equation}
r_{\mathrm{coll\_drone}} 
= \sum_{j \in \mathcal{N}_i}
\frac{k_2}{d_{ij}^2 + k_3},
\end{equation}
where $d_{ij}$ denotes the distance between agent $i$ and neighboring drone $j$ within its FOV, 
$k_2 = -400$, and $k_3 = 50$ are scaling constants. 
This increases the penalty as agents approach each other, encouraging spatial separation.

Finally, a flag mechanism switches between obstacle-aware and metric-based evaluation. When $flag=0$, the algorithm prioritizes obstacle avoidance and feasibility screening. When $flag=1$, Euclidean and wind-adjusted distances are fused to form a synthetic metric used for path chronologizing and evaluation. The algorithm maintains the remaining budget $B$, current vertex $v$, a delivery flag $\textsf{del}$, and the executed path $P$ represents the termination status.

\subsection{Trajectory Optimization}

Module~4 of Fig. \ref{fig:framework} translates discrete waypoints into dynamically feasible trajectories. Straight-line interpolation is insufficient due to heading continuity and minimum turning radius constraints. Therefore, Dubins curves are used to generate curvature-bounded paths.

An obstacle detection stage first filters geometrically infeasible segments. Valid segments and states $S$ are then refined via Dubins path synthesis, producing smooth trajectories parameterized by turning angle $\varphi$ and path length $D$. The resulting trajectory length contributes to energy estimation in the BER metric map.

Fig.~\ref{fig:framework} illustrates the proposed modular framework for robust energy-aware UAV delivery under wind uncertainty. The overall objective is not merely to minimize energy consumption, but to determine and execute feasible delivery missions when future wind conditions are unknown and energy costs evolve online.

\section{Baseline Methods}
We compare BER against three canonical online routing policies that share the same wind sensitivity parameter $\lambda$. All methods are evaluated on the same underlying graph topology, defined in Eq. \ref{eq:edge}. SER uses the initial cost snapshot $c_0$, while RER/GER/BER use online-updated costs $c_t$.

\subsubsection{Static Energy-Optimal Routing (SER)} It serves as a conservative baseline that calculates a complete delivery route at the beginning of the mission, under the assumption that the initial wind conditions remain constant. 

\subsubsection{Online Replanning with Updated Edge Costs (RER)} RER dynamically updates the planned route in response to evolving wind conditions. 
At each visited vertex, the UAV observes the local wind and updates the current wind estimate. 
The remaining path is then recomputed using the updated edge costs.
each edge $e$ is assigned a risk-aware surrogate cost
\begin{equation}
c_t(e)=EnergyCost(e,\hat{w}_t)+\lambda\,Unc(e,\hat{w}_t),
\end{equation}
where $\textsc{EnergyCost}(\cdot)$ denotes the predicted energy consumption under the current wind estimate $\hat{w}_t$, and $\textsc{Unc}(\cdot)$ captures wind-induced uncertainty. 
The parameter $\lambda$ controls the trade-off between energy efficiency and uncertainty-aware risk.

\subsubsection{Greedy Energy-Minimizing Routing (GER)} At each decision point, GER selects the outgoing edge that incurs the lowest instantaneous energy cost under the current wind conditions, without regard for long-term implications or mission completion.
From the current vertex  \(v_t\) , GER identifies the optimal edge as follows:

\begin{equation} e^\star = \arg\min_{e \in \mathcal{E}(v_t)} c_t(e), \end{equation}
where  \(\mathcal{E}(v_t\))  represents the set of outgoing edges from  \(v_t\) . The UAV then traverses edge  \(e^\star\)  and repeats this process at the next vertex.

\section{Experiments}

\subsection{Setup}

Our simulation experiments for all algorithms and setups were conducted on a computer equipped with AMD Ryzen 7 5800H CPU and NVIDIA RTX 3050 GPU.
We evaluate the proposed routing framework on two types of environments:

(i) \emph{Synthetic ER graphs}: To generate controllable yet realistic delivery environments, we construct high-fidelity simulation scenes in Unreal Engine~4 (UE4) \cite{sanders2016introduction}, which provides physically-based lighting and wind field modeling. The simulated urban environments capture complex illumination and wind interactions around buildings, allowing realistic aerodynamic disturbances to be reproduced. Based on these environments, delivery scenarios are abstracted into Erd\H{o}s--R\'enyi (ER) graphs, where vertices represent waypoints or customer locations (including the depot) and edges denote feasible flight segments. Each edge is associated with a time-varying energy cost induced by the wind conditions observed along that segment.

(ii) \emph{Quasi-real wind logs}: To evaluate routing performance under realistic wind dynamics, we replay wind measurements from a publicly available truck--UAV delivery dataset\cite{rigoni2022delivery}. The dataset contains time-indexed wind observations collected during simulated delivery operations, including variations in wind speed and direction over time. These logs are streamed during simulation to update edge costs online, preserving realistic temporal fluctuations of environmental conditions while maintaining a controllable graph topology~\cite{rigoni2022delivery}.

Wind uncertainty is discretized into either \(4\) or \(8\) classes, which determines how continuous wind observations are mapped to discrete wind states used by the planner.
We conduct 10 rounds of experiments for our method as well as all
mentioned baselines. Within each round, we vary the number
of customers and drones of simulation environment. Each drone is assumed to share identical flight dynamics, cruising speed, and battery endurance

Each mission outcome is categorized into four mutually exclusive events:
\emph{Success (SUC)} indicates the UAV completes the route and returns safely within budget;
\emph{Delivered (DEL)} indicates the payload is delivered but the mission does not fully satisfy the return feasibility criterion;
\emph{Fail (FAIL)} indicates infeasible execution (e.g., energy depletion or violation of hard constraints);
\emph{Aborted (ABRT)} indicates early termination due to online feasibility checks triggering a safe abort.

\subsection{Results}

\begin{figure}
    \centering
    \includegraphics[width=1\linewidth]{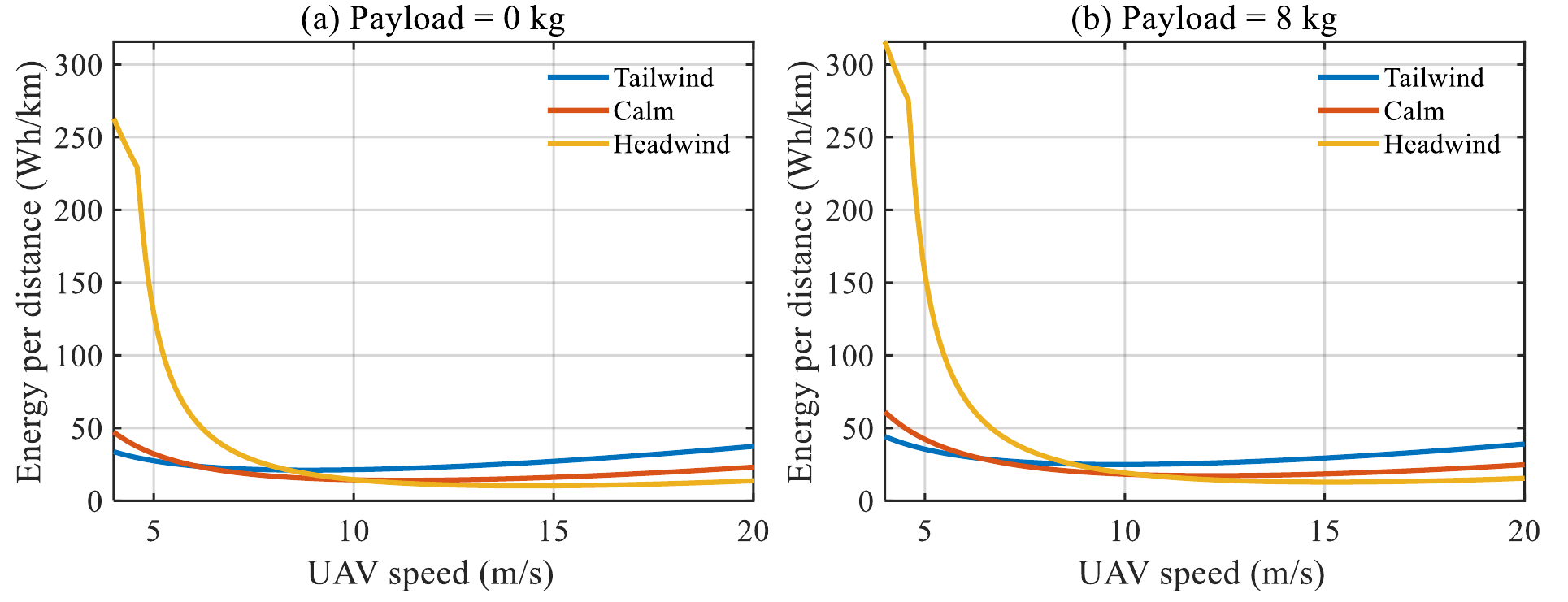}
    \caption{Energy per distance (Wh/km) modeling as a function of UAV speed under different wind conditions and payload states. 
(a) No payload (0 kg). 
(b) Loaded condition (8 kg). 
}
    \label{fig:energy}
\end{figure}

Fig. \ref{fig:energy} illustrates the relationship between UAV speed and energy consumption per distance under different wind conditions and payloads. Headwind significantly increases energy demand, particularly at low speeds, while tailwind slightly reduces consumption. In both payload settings, an intermediate speed range minimizes energy per distance, indicating an optimal cruise speed for energy-efficient flight.

\begin{table*}[t]
\centering
\caption{Quasi-real mission outcomes under different wind discretizations.
Results are reported as mean $\pm$ standard deviation (\%). 
Metrics include Success rate (SUC), Delivery-only (DEL), Failure (FAIL), and Abort (ABRT).}
\label{tab:quasi_real_comparison_final}
\renewcommand{\arraystretch}{1.15}
\setlength{\tabcolsep}{3.5pt}
\small

\resizebox{\textwidth}{!}{%
\begin{tabular}{cc|cccc|cccc|cccc|cccc}
\toprule
\multirow{2}{*}{Graph} & \multirow{2}{*}{B} 
& \multicolumn{4}{c|}{\textbf{SER}}
& \multicolumn{4}{c|}{\textbf{RER}}
& \multicolumn{4}{c|}{\textbf{GER}}
& \multicolumn{4}{c}{\textbf{BER (Ours)}} \\
\cmidrule(lr){3-6}
\cmidrule(lr){7-10}
\cmidrule(lr){11-14}
\cmidrule(lr){15-18}
 &  & SUC & DEL & FAIL & ABRT
    & SUC & DEL & FAIL & ABRT
    & SUC & DEL & FAIL & ABRT
    & SUC & DEL & FAIL & ABRT \\
\midrule

\multicolumn{18}{c}{\textbf{4 Wind Classes}} \\
\midrule

ER & 100
& 89.6$\pm$1.2 & 1.1$\pm$0.4 & 0.0$\pm$0.0 & 9.3$\pm$1.3
& 90.3$\pm$2.0 & 9.7$\pm$1.9 & 0.0$\pm$0.0 & 0.0$\pm$0.0
& 8.1$\pm$3.4 & 24.2$\pm$3.1 & 67.7$\pm$4.2 & 0.0$\pm$0.0
& \textbf{93.8$\pm$1.1} & 2.4$\pm$0.8 & 0.0$\pm$0.0 & 3.8$\pm$0.9 \\

ER & 50
& 55.2$\pm$1.6 & 0.0$\pm$0.0 & 0.0$\pm$0.0 & 44.8$\pm$1.7
& 56.4$\pm$2.5 & 38.2$\pm$2.3 & 5.4$\pm$1.6 & 0.0$\pm$0.0
& 3.6$\pm$1.9 & 20.3$\pm$2.6 & 76.1$\pm$3.8 & 0.0$\pm$0.0
& \textbf{61.5$\pm$1.3} & 4.2$\pm$1.1 & 0.0$\pm$0.0 & 34.3$\pm$1.5 \\

\midrule

\multicolumn{18}{c}{\textbf{8 Wind Classes}} \\
\midrule

ER & 100
& 90.8$\pm$1.1 & 0.9$\pm$0.3 & 0.0$\pm$0.0 & 8.3$\pm$1.2
& 91.7$\pm$1.8 & 8.3$\pm$1.7 & 0.0$\pm$0.0 & 0.0$\pm$0.0
& 7.4$\pm$3.1 & 23.1$\pm$3.0 & 69.5$\pm$4.0 & 0.0$\pm$0.0
& \textbf{95.6$\pm$0.9} & 1.8$\pm$0.6 & 0.0$\pm$0.0 & 2.6$\pm$0.7 \\

ER & 50
& 57.6$\pm$1.5 & 0.0$\pm$0.0 & 0.0$\pm$0.0 & 42.4$\pm$1.6
& 57.9$\pm$2.3 & 40.0$\pm$2.2 & 2.1$\pm$0.9 & 0.0$\pm$0.0
& 3.4$\pm$1.7 & 21.0$\pm$2.5 & 75.6$\pm$3.6 & 0.0$\pm$0.0
& \textbf{66.2$\pm$1.2} & 3.6$\pm$0.9 & 0.0$\pm$0.0 & 30.2$\pm$1.3 \\

\bottomrule
\end{tabular}%
}

\vspace{3pt}

\end{table*}

\begin{table*}[t]
\centering
\small
\setlength{\tabcolsep}{6pt}
\renewcommand{\arraystretch}{1.15}

\caption{\textbf{Ablation study of BER.} We remove \emph{Budget Gate}, \emph{Wind costs} (), \emph{Risk term} (uncertainty-aware penalty), and \emph{Trajectory Optimization} respectively at one time. We also evaluate the role of semantic clustering compared to traditional K-Means clustering \cite{pan2019uav}.
Metrics are reported as mean $\pm$ std over $10$ trials. Better results are indicated with arrows.
}
\label{tab:ablation_ber}

\begin{tabularx}{\linewidth}{l *{8}{>{\centering\arraybackslash}X}}
\toprule
\textbf{Variant} 
& \makecell{\textbf{SUC} $\uparrow$\\(\%)} 
& \makecell{\textbf{ABRT} $\downarrow$\\(\%)} 
& \makecell{\textbf{FAIL} $\downarrow$\\(\%)} 
& \makecell{\textbf{Energy} $\downarrow$\\(Wh)} 
& \makecell{\textbf{Margin} $\uparrow$\\(Wh)} 
& \makecell{\textbf{Time} $\downarrow$\\(s)} 
& \makecell{\textbf{Max Turn} $\downarrow$\\(deg)} 
 \\
\midrule
\textbf{BER (Full)} 
& \textbf{92.4 $\pm$ 2.8}  & \underline{4.1 $\pm$ 1.7} 
& \textbf{3.5 $\pm$ 1.6} 
& \textbf{95.2 $\pm$ 4.9} 
& \textbf{12.6 $\pm$ 3.4} 
& 148.7 $\pm$ 9.6 
& \textbf{11.3 $\pm$ 2.1}  \\
\midrule
w/o \textbf{Budget Gate} 
& 74.6 $\pm$ 6.1 
& 6.3 $\pm$ 2.9 
& 19.1 $\pm$ 5.8 
& 110.4 $\pm$ 6.7 
& 1.9 $\pm$ 3.1 
& 162.8 $\pm$ 13.4 
& 12.7 $\pm$ 3.0 \\
w/o \textbf{LLM} 
& 86.1 $\pm$ 4.7 & 5.5 $\pm$ 2.3 & 8.4 $\pm$ 3.1 & 101.3 $\pm$ 6.1 & 7.8 $\pm$ 3.7 & 154.8 $\pm$ 11.7 & 11.8 $\pm$ 2.5 \\
w \textbf{K-Means} 
& 88.2 $\pm$ 4.0 & 5.0 $\pm$ 2.1 & 6.8 $\pm$ 2.8 & 99.6 $\pm$ 5.7 & 8.9 $\pm$ 3.5 & 151.9 $\pm$ 10.6 & 11.6 $\pm$ 2.4 \\
w/o \textbf{Wind} 
& 82.7 $\pm$ 4.4 
& 9.8 $\pm$ 3.1 
& 7.5 $\pm$ 2.6 
& 106.8 $\pm$ 5.8 
& 6.4 $\pm$ 3.0 
& 158.6 $\pm$ 11.2 
& 11.9 $\pm$ 2.6  \\
w/o \textbf{Risk} 
& 88.9 $\pm$ 4.2 
& 4.7 $\pm$ 2.0 
& 6.4 $\pm$ 3.9 
& \underline{96.7 $\pm$ 6.8 }
& 9.1 $\pm$ 4.6 
& \underline{146.9 $\pm$ 10.8 }
& \underline{11.5 $\pm$ 2.3}   \\
w/o \textbf{Opt} 
& \underline{90.8 $\pm$ 3.2} 
& \textbf{3.9 $\pm$ 1.8 }
& \underline{5.3 $\pm$ 2.5 }
& 97.5 $\pm$ 5.2 
& \underline{10.8 $\pm$ 3.8 }
& \textbf{145.2 $\pm$ 9.9 }
& 104.6 $\pm$ 14.8   \\

\bottomrule
\end{tabularx}

\vspace{2pt}

\end{table*}

\begin{figure}
    \centering
    \includegraphics[width=1\linewidth]{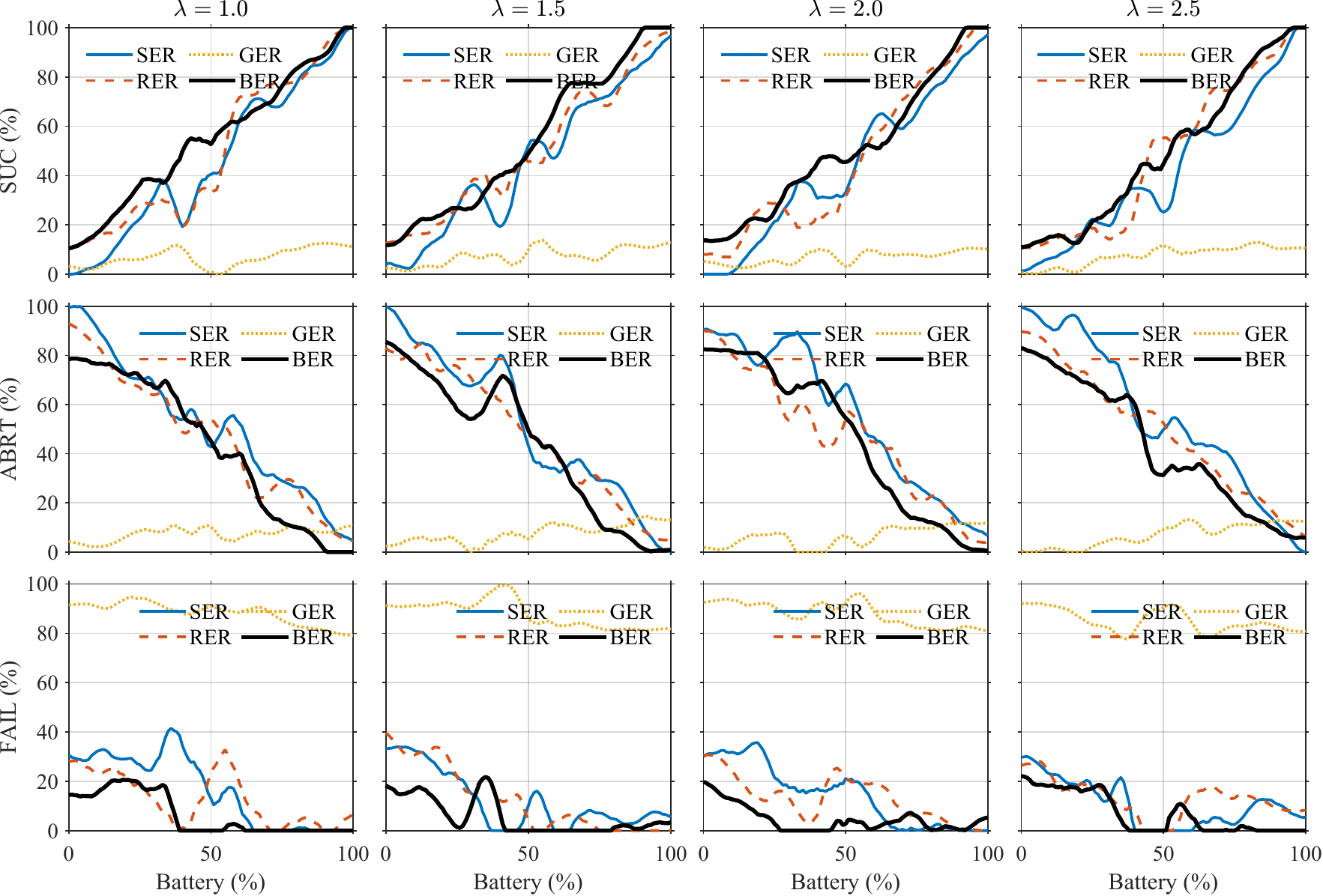}
    \caption{Comparative performance of SER, RER, GER, and BER under varying risk sensitivity (\(\lambda\) = 1.0–2.5). The plots illustrate how success (SUC), abort (ABRT), and failure (FAIL) probabilities evolve with the remaining battery level during mission execution, highlighting BER’s robustness and stability across conditions}
    \label{fig:sensitivity}
\end{figure}

Fig. \ref{fig:sensitivity} illustrates the performance trends of different routing strategies under varying risk sensitivity parameters (\(\lambda\)). Across battery levels, BER generally maintains comparable or higher success rates while exhibiting lower failure rates than SER, RER, and GER. The results indicate that BER provides relatively stable behavior under different risk preferences and energy conditions.

As shown in Tab. \ref{tab:quasi_real_comparison_final}, while SER emphasizes safety and predictability, it is unable to take advantage of favorable changes in wind conditions or to recover from unexpected increases in energy expenditure.
RER is more proactive than SER and can salvage missions that would otherwise fail. However, this comes at the expense of increased computational overhead and heightened sensitivity to transient wind fluctuations.
GER requires no global planning and incurs minimal computational cost. However, by disregarding long-term consequences, it frequently leads to dead ends or energy depletion before mission completion. Under both coarse (4-class) and finer (8-class) wind discretizations, BER improves SUC by approximately 3–8\% over SER and RER, with the advantage becoming more pronounced in low-budget scenarios (B=50). Notably, when battery capacity is limited, BER retains a 66.2\% success rate under 8 wind classes, outperforming SER and RER by nearly 9 percentage points. In contrast, GER exhibits a high failure rate (over 67\%), indicating that purely greedy energy minimization is insufficient under wind uncertainty.

\subsection{Ablation Study}
Our ablation experiments are conducted on the same wind-field log dataset to ensure consistent environmental conditions across variants~\cite{rigoni2022delivery}. As shown in Table \ref{tab:ablation_ber}, removing the Budget Gate leads to the largest degradation in mission feasibility, reducing the success rate from 92.4\% to 74.6\% while significantly increasing failures, indicating that energy-feasibility checking is critical for safe routing. Disabling wind-awareness increases both abort and failure rates and results in higher energy consumption, highlighting the importance of modeling wind-conditioned costs. Removing the risk term slightly reduces robustness, reflected by increased failure variance. Finally, removing trajectory optimization drastically increases the maximum turning angle, confirming its role in improving trajectory executability without significantly affecting mission success.

\section{Discussion and Limitations}
The experimental results reveal the significance of adaptive routing in the face of wind uncertainty. While static routing strategies like SER perform adequately under stable wind conditions, they falter as conditions change due to their inability to update energy estimates in real-time. Conversely, purely reactive strategies such as GER often fail, as minimizing instantaneous energy costs does not ensure mission success. The proposed BER framework consistently yields higher success rates across various wind scenarios and battery budgets, demonstrating the importance of incorporating return-feasibility checks and online energy updates for safe UAV delivery in dynamic environments.

Ablation studies further elucidate the impact of each module. The removal of the battery budget leads to the greatest performance decline, highlighting the crucial role of energy-feasibility verification in ensuring mission safety. Disabling wind-sensitive cost modeling results in increased abort and failure rates, indicating that overlooking wind-related energy variations causes systematic planning errors. In contrast, eliminating the risk term only slightly affects robustness, while trajectory optimization primarily influences path executability rather than overall mission success. Collectively, these findings affirm that effective UAV delivery in wind-affected environments necessitates a cohesive integration of energy budgeting, environmental awareness, and adaptive routing.

Despite these promising results, several limitations remain. 
First, the evaluation is evaluated in simulation environments with quasi-real wind logs. 
Second, wind dynamics are discretized into a limited number of directional classes, which may not fully capture fine-grained turbulence structures commonly observed in dense urban airflow.

\section{Conclusion}
This work investigates energy-aware routing for truck-UAV delivery missions operating under uncertain wind conditions. We model the delivery environment as a time-dependent energy graph in which edge costs evolve according to observed wind dynamics. Based on this formulation, we propose an online planning framework that integrates wind perception, residual energy budgeting, and risk-sensitive routing decisions. Experimental results on synthetic delivery graphs and quasi-real wind logs show that BER consistently improves mission success rates while maintaining stable behavior across different battery budgets and wind discretizations.

The results demonstrate that incorporating feasibility-aware energy constraints significantly enhances mission reliability compared with static or greedy routing strategies. Beyond the proposed algorithm, this study highlights the broader importance of integrating environmental awareness into energy-constrained aerial planning. Future work will investigate scalable planning strategies and higher-fidelity 3D wind modeling to further improve real-world applicability.

\bibliographystyle{IEEEtran}

\bibliography{references}

@article{han2020energy,
  title={Energy-efficient UAV communications under stochastic trajectory: A Markov decision process approach},
  author={Han, Di and Chen, Wei and Liu, Jianqing},
  journal={IEEE Transactions on Green Communications and Networking},
  volume={5},
  number={1},
  pages={106--118},
  year={2020},
  publisher={IEEE}
}

@article{michel2023energy,
  title={Energy-optimal unmanned aerial vehicles motion planning and control based on integrated system physical dynamics},
  author={Michel, Nicolas and Wei, Peng and Kong, Zhaodan and Lin, Xinfan},
  journal={Journal of Dynamic Systems, Measurement, and Control},
  volume={145},
  number={4},
  pages={041002},
  year={2023},
  publisher={American Society of Mechanical Engineers}
}

@article{mellinger2012trajectory,
  title={Trajectory generation and control for precise aggressive maneuvers with quadrotors},
  author={Mellinger, Daniel and Michael, Nathan and Kumar, Vijay},
  journal={The International Journal of Robotics Research},
  volume={31},
  number={5},
  pages={664--674},
  year={2012},
  publisher={SAGE Publications Sage UK: London, England}
}

@article{dorling2016vehicle,
  title={Vehicle routing problems for drone delivery},
  author={Dorling, Kevin and Heinrichs, Jordan and Messier, Geoffrey G and Magierowski, Sebastian},
  journal={IEEE Transactions on Systems, Man, and Cybernetics: Systems},
  volume={47},
  number={1},
  pages={70--85},
  year={2016},
  publisher={IEEE}
}

@inproceedings{rigoni2022delivery,
  title={Delivery with UAVs: a simulated dataset via ATS},
  author={Rigoni, Giulio and Pinotti, Cristina M and Bhumika and Das, Debasis and Das, Sajal K},
  booktitle={2022 IEEE 95th Vehicular Technology Conference:(VTC2022-Spring)},
  pages={1--6},
  year={2022},
  organization={IEEE}
}

@book{sanders2016introduction,
  title={An introduction to Unreal engine 4},
  author={Sanders, Andrew},
  year={2016},
  publisher={AK Peters/CRC Press}
}

@article{wang2019overview,
  title={An overview of various kinds of wind effects on unmanned aerial vehicle},
  author={Wang, Bo Hang and Wang, Dao Bo and Ali, Zain Anwar and Ting Ting, Bai and Wang, Hao},
  journal={Measurement and Control},
  volume={52},
  number={7-8},
  pages={731--739},
  year={2019},
  publisher={SAGE Publications Sage UK: London, England}
}

@article{chiang2019impact,
  title={Impact of drone delivery on sustainability and cost: Realizing the UAV potential through vehicle routing optimization},
  author={Chiang, Wen-Chyuan and Li, Yuyu and Shang, Jennifer and Urban, Timothy L},
  journal={Applied energy},
  volume={242},
  pages={1164--1175},
  year={2019},
  publisher={Elsevier}
}

@article{wang2024review,
  title={A review of research on shortest path planning algorithms for mobile robots},
  author={Wang, Lutong and Yang, Xiaodong and Liu, Shouguo and Liu, Hongbo and Peng, Jiakuan and Liu, Weifeng},
  journal={Recent Patents on Engineering},
  year={2024},
  publisher={Bentham Science Publishers}
}

@article{zaki2016comprehensive,
  title={Comprehensive survey on dynamic graph models},
  author={Zaki, Aya and Attia, Mahmoud and Hegazy, Doaa and Amin, Safaa},
  journal={International Journal of Advanced Computer Science and Applications},
  volume={7},
  number={2},
  year={2016},
  publisher={Science and Information (SAI) Organization Limited}
}

@article{rasmussen2008tree,
  title={Tree search algorithm for assigning cooperating UAVs to multiple tasks},
  author={Rasmussen, Steven J and Shima, Tal},
  journal={International Journal of Robust and Nonlinear Control: IFAC-Affiliated Journal},
  volume={18},
  number={2},
  pages={135--153},
  year={2008},
  publisher={Wiley Online Library}
}

@article{du2023efficient,
  title={Efficient tree-svd for subset node embedding over large dynamic graphs},
  author={Du, Xinyu and Zhang, Xingyi and Wang, Sibo and Huang, Zengfeng},
  journal={Proceedings of the ACM on Management of Data},
  volume={1},
  number={1},
  pages={1--26},
  year={2023},
  publisher={ACM New York, NY, USA}
}

@article{2024Energy,
  title={Energy-Optimized Path Planning for Uas in Varying Winds Via Reinforcement Learning},
  author={ Banerjee, Portia  and  Bradner, Kevin },
  journal={AIAA AVIATION FORUM AND ASCEND 2024},
  year={2024},
}

@article{ferone2017shortest,
  title={Shortest paths on dynamic graphs: a survey},
  author={Ferone, Daniele and Festa, Paola and Napoletano, Antonio and Pastore, Tommaso},
  journal={Pesquisa Operacional},
  volume={37},
  number={3},
  pages={487--508},
  year={2017},
  publisher={SciELO Brasil}
}

@inproceedings{yin2021learning,
  title={Learning shortest paths on large dynamic graphs},
  author={Yin, Jiaming and Rao, Weixiong and Zhang, Chenxi},
  booktitle={2021 22nd IEEE International Conference on Mobile Data Management (MDM)},
  pages={201--208},
  year={2021},
  organization={IEEE}
}

@article{zhu2021uav,
  title={UAV trajectory planning in wireless sensor networks for energy consumption minimization by deep reinforcement learning},
  author={Zhu, Botao and Bedeer, Ebrahim and Nguyen, Ha H and Barton, Robert and Henry, Jerome},
  journal={IEEE Transactions on Vehicular Technology},
  volume={70},
  number={9},
  pages={9540--9554},
  year={2021},
  publisher={IEEE}
}

@article{Wu2021ReinforcementLB,
  title={Reinforcement Learning Based Truck-and-Drone Coordinated Delivery},
  author={Guohua Wu and Mingfeng Fan and Jianmai Shi and Yanghe Feng},
  journal={IEEE Transactions on Artificial Intelligence},
  year={2021},
  volume={4},
  pages={754-763},
  url={https://api.semanticscholar.org/CorpusID:237814941}
}

@article{Bi2024TruckDroneDO,
  title={Truck-Drone Delivery Optimization Based on Multi-Agent Reinforcement Learning},
  author={Zhiliang Bi and Xiwang Guo and Jiacun Wang and Shujin Qin and Guanjun Liu},
  journal={Drones},
  year={2024},
  url={https://api.semanticscholar.org/CorpusID:267103096}
}

@inproceedings{duan2024energy,
  title={Energy-optimized planning in non-uniform wind fields with fixed-wing aerial vehicles},
  author={Duan, Yufei and Achermann, Florian and Lim, Jaeyoung and Siegwart, Roland},
  booktitle={2024 IEEE/RSJ International Conference on Intelligent Robots and Systems (IROS)},
  pages={3116--3122},
  year={2024},
  organization={IEEE}
}

@inproceedings{Xu2025UAVAT,
  title={UAV assisted truck delivery route optimization via reinforcement learning},
  author={Yawen Xu and Hongwei Ma and Dequn Zhao and Qianhua Deng and Zikai Liu},
  booktitle={International Conference on Algorithms, Microchips and Network Applications},
  year={2025},
  url={https://api.semanticscholar.org/CorpusID:278731744}
}

@article{Hu2025MultiDroneTruckCD,
  title={Multi-Drone-Truck Collaborative Delivery with En Route Operations: A Hierarchical MARL-Based Approach},
  author={Shunong Hu and Bing Li and Rongqing Zhang},
  journal={2025 IEEE International Conference on Robotics and Automation (ICRA)},
  year={2025},
  pages={12993-12999},
  url={https://api.semanticscholar.org/CorpusID:281093768}
}

@article{Park2024LearningBasedCM,
  title={Learning-Based Cooperative Mobility Control for Autonomous Drone-Delivery},
  author={Soohyun Park and Chan Yi Park and Joongheon Kim},
  journal={IEEE Transactions on Vehicular Technology},
  year={2024},
  volume={73},
  pages={4870-4885},
  url={https://api.semanticscholar.org/CorpusID:265164677}
}

@INPROCEEDINGS{4434966,
  author={Chitsaz, Hamidreza and LaValle, Steven M.},
  booktitle={2007 46th IEEE Conference on Decision and Control}, 
  title={Time-optimal paths for a Dubins airplane}, 
  year={2007},
  volume={},
  number={},
  pages={2379-2384},
  doi={10.1109/CDC.2007.4434966}}

@article{2025Multi,
  title={Multi-Agent Reinforcement Learning for truck–drone routing in smart logistics: A comprehensive review},
  author={ Arishi, Ali  and  Ahuja, Paras },
  journal={Computers Electrical Engineering},
  volume={127},
  number={PartA},
  year={2025},
}

@article{Murray2015TheFS,
  title={The flying sidekick traveling salesman problem: Optimization of drone-assisted parcel delivery},
  author={Chase C. Murray and Amanda Chu},
  journal={Transportation Research Part C-emerging Technologies},
  year={2015},
  volume={54},
  pages={86-109},
  url={https://api.semanticscholar.org/CorpusID:8846258}
}

@article{Pasha2026EnablingTM,
  title={Enabling the multi-LR ability of drones in the multi-visit truck-drone routing problem with pickup and delivery},
  author={S. Ameer Pasha and S.Mehdi Sajadifar},
  journal={Transportation Research Part E: Logistics and Transportation Review},
  year={2026},
  url={https://api.semanticscholar.org/CorpusID:285250958}
}

@ARTICLE{9766183,
  author={Kong, Fanhui and Li, Jianqiang and Jiang, Bin and Wang, Huihui and Song, Houbing},
  journal={IEEE Transactions on Intelligent Transportation Systems}, 
  title={Trajectory Optimization for Drone Logistics Delivery via Attention-Based Pointer Network}, 
  year={2023},
  volume={24},
  number={4},
  pages={4519-4531},
  doi={10.1109/TITS.2022.3168987}}

@article{Lauri_2023,
   title={Partially Observable Markov Decision Processes in Robotics: A Survey},
   volume={39},
   ISSN={1941-0468},
   url={http://dx.doi.org/10.1109/TRO.2022.3200138},
   DOI={10.1109/tro.2022.3200138},
   number={1},
   journal={IEEE Transactions on Robotics},
   publisher={Institute of Electrical and Electronics Engineers (IEEE)},
   author={Lauri, Mikko and Hsu, David and Pajarinen, Joni},
   year={2023},
   month=feb, pages={21–40} }

@article{chodnicki2022energy,
  title={Energy efficient UAV flight control method in an environment with obstacles and gusts of wind},
  author={Chodnicki, Marcin and Siemiatkowska, Barbara and Stecz, Wojciech},
  journal={Energies},
  volume={15},
  number={10},
  pages={3730},
  year={2022},
  publisher={MDPI}
}

@article{tong2022optimal,
  title={Optimal route planning for truck--drone delivery using variable neighborhood tabu search algorithm},
  author={Tong, Bao and Wang, Jianwei and Wang, Xue and Zhou, Feihao and Mao, Xinhua and Zheng, Wenlong},
  journal={Applied sciences},
  volume={12},
  number={1},
  pages={529},
  year={2022},
  publisher={MDPI}
}

@article{meier2022wind,
  title={Wind estimation with multirotor UAVs},
  author={Meier, Kilian and Hann, Richard and Skaloud, Jan and Garreau, Arthur},
  journal={Atmosphere},
  volume={13},
  number={4},
  pages={551},
  year={2022},
  publisher={MDPI}
}

@ARTICLE{11214461,
  author={Xing, Xiaojun and Ma, Yuanqiang and Lei, Yichen and Li, Yan and Xiao, Bing},
  journal={IEEE Transactions on Vehicular Technology}, 
  title={Multi-UAV Rendezvous Trajectory Planning Based on Improved MADDPG Algorithm in Complex Dynamic Obstacle Environments}, 
  year={2025},
  volume={},
  number={},
  pages={1-12},
  doi={10.1109/TVT.2025.3624052}}

@article{Deng2025TargetAA,
  title={Target Allocation and Air–Ground Coordination for UAV Cluster Airspace Security Defense},
  author={Chang-shou Deng and Xi Fang},
  journal={Drones},
  year={2025},
  url={https://api.semanticscholar.org/CorpusID:282978382}
}

@inproceedings{pan2019uav,
  title={Uav delivery planning based on k-means++ clustering and genetic algorithm},
  author={Pan, Sinuo},
  booktitle={2019 5th International Conference on Control Science and Systems Engineering (ICCSSE)},
  pages={14--18},
  year={2019},
  organization={IEEE}
}

\end{document}